\documentclass[final,12pt]{msml2022} 

\title{Error-in-variables modelling for operator learning}
\usepackage{times}

\usepackage{bm}
\usepackage{epstopdf}

\usepackage{mathtools}
\mathtoolsset{showonlyrefs}

\newcommand{\sym}[1]{(\protect\includegraphics[height=6pt]{styles/#1.pdf})}

\msmlauthor{%
 \Name{Ravi G. Patel} \Email{rgpatel@sandia.gov}\\
 \addr Center for Computing Research, Sandia National Laboratories
 \AND
 \Name{Indu Manickam} \Email{imanick@sandia.gov}\\
 \addr Applied Information Sciences, Sandia National Laboratories
 \AND
 \Name{Myoungkyu Lee} \Email{mklee@ua.edu}\\
 \addr  Aerospace Engineering and Mechanics, University of Alabama
 \AND
 \Name{Mamikon Gulian} \Email{mgulian@sandia.gov}\\
 \addr Quantitative Modeling and Analysis, Sandia National Laboratories%
}

\makeatletter
 \let\Ginclude@graphics\@org@Ginclude@graphics
\makeatother

\begin{document}

\maketitle

\begin{abstract}%
Deep operator learning has emerged as a promising tool for reduced-order modelling and PDE model discovery. Leveraging the expressive power of deep neural networks, especially in high dimensions, such methods learn the mapping between functional state variables. While proposed methods have assumed noise only in the dependent variables, experimental and numerical data for operator learning typically exhibit noise in the independent variables as well, since both variables represent signals that are subject to measurement error.  In regression on scalar data, failure to account for noisy independent variables can lead to biased parameter estimates. With noisy independent variables, linear models fitted via ordinary least squares (OLS) will show attenuation bias, wherein the slope will be underestimated. In this work, we derive an analogue of attenuation bias for linear operator regression with white noise in both the independent and dependent variables, showing that the norm upper bound of the operator learned via OLS decreases with increasing noise in the independent variable. In the nonlinear setting, we computationally demonstrate underprediction of the action of the Burgers operator in the presence of noise in the independent variable. We propose error-in-variables (EiV) models for two operator regression methods, MOR-Physics and DeepONet, and demonstrate that these new models reduce bias in the presence of noisy independent variables for a variety of operator learning problems. Considering the Burgers operator in 1D and 2D, we demonstrate that EiV operator learning robustly recovers operators in high-noise regimes that defeat OLS operator learning. We also introduce an EiV model for time-evolving PDE discovery and show that OLS and EiV perform similarly in learning the Kuramoto-Sivashinsky evolution operator from corrupted data, suggesting that the effect of bias in OLS operator learning depends on the regularity of the target operator.
\end{abstract}


\section{Introduction}
Operator regression, or operator learning, has emerged as an important field in scientific computing that focuses on the flexible and expressive discovery of partial differential equations (PDEs) and related models. These methods fit parameterizations of operators, i.e., mappings between functions, using observations of the input and output of the operators. Recently, operator regression methods have begun to incorporate techniques from machine learning such as deep neural networks and Gaussian processes, opening the way for novel applications in scientific computing challenges, such as data-driven model discovery \citep{trask2019gmlsnets}, surrogate models of data-to-solution maps \citep{you2022nonlocal, cai2021deepm}, and closure models in fluid flow \citep{duraisamy2019turbulence}.
However, current operator regression methods based on deep learning assume that the independent variable, i.e. the input to the operator, is free of noise. The objective of this work is to explore the bias in models learned from noisy independent variables and to propose a correction, applicable to a wide variety of operator regression architectures, based on error-in-variables methods in classical statistics.

Fitting parameterized operators to observations is a classical problem and has been approached, e.g., with  Bayesian methods \citep{pang2017discovering,stuart2010inverse,trillos2017bayesian}, PDE-constrained optimization \citep{d2016identification,burkovska2021optimization}, and physics-informed Gaussian processes \citep{gulian2019machine,raissi2017machine}. Earlier works leveraged significant prior knowledge of physics and fitted small numbers of physically interpretable parameters.
In contrast, recent methods based on high-dimensional deep neural network (DNN) parameterization of operators have established deep operator learning as a widely applicable and domain-agnostic area of scientific computing.
Deep operator learning methods vary based on the discretizations of function spaces inherent in each method as well as the architecture and training of the DNN. They include modal methods \citep{patel2018nonlinear,patel2021physics,qin2019data,li2021fourier}, graph based methods \citep{anandkumar2020neural,li2020neural}, PCA based methods \citep{bhattacharya2020model}, meshless methods \citep{trask2019gmlsnets}, trunk-branch based methods \citep{lu2019deeponet,cai2021deepm}, and time-stepping methods \citep{You2021,Long2018,qin2019data}. Such methods can be purely data-driven or incorporate knowledge from physics \citep{wang2021learning}, and can be utilized as general operator surrogates or be specifically for PDE model discovery \citep{patel2021physics}. 
Similar techniques in the computer science and statistics literature have been described as function-to-function regression within the field of functional data analysis (FDA) \citep{Ramsay2005} but applied outside the PDE context. These methods include reproducing kernel Hilbert space approaches \citep{Yuan2010,Kadri2016}, additive models \citep{kim2018}, wavelet-based approaches \citep{meyer2015}, and neural network models \citep{Rao2021,Kou2019}. See \cite{Morris2015} for a broad overview of the topic.

The effect of noisy inputs in the training and test data for deep operator regression methods is relatively unexplored, despite being widely studied in the related context of adversarial examples in deep learning \citep{szegedy2013intriguing, yoshida2017spectral}.
In single and multivariate statistics, error in the independent variables leads to inconsistent parameter estimates without proper error modelling. In particular, linear models fit via ordinary least squares (OLS) to scalar data $\{(x+\epsilon, y)\}$ with error in the independent variable persistently underpredict the slope as
\begin{equation} \label{eq:attenuation}
    m^* = m \frac{\mathrm{var}(x)}{\mathrm{var}(x) + \mathrm{var}(\epsilon)},
\end{equation}
where $m^*$ is the predicted slope, $m$ is the true slope, $x$ is the independent variable, and $\epsilon$ is the error in $x$ \citep{Hutcheon2010}. This phenomenon is known as \emph{attenuation bias} and methods such as total least squares \citep{Markovsky2007}, and Deming regression \citep{linnet1993evaluation} have been developed to counteract it. More generally, including in nonlinear contexts, an error-in-variables (EiV) model is required to correct for inconsistency arising from error in the independent variables. Typically, EiV models are highly specialized to specific problems. See \cite{zwanzig2000estimation,Chen2011} for overviews of this topic. 

For multivariate linear regression, the method of \cite{xu2007covariate} can recover models given independent variables corrupted by additive noise. The Compensated Matrix Uncertainty selector method  \citep{rosenbaum2013improved} has been developed to recover over-parameterized models under the assumption that the parameter vector is sparse and the variance of the noise can be estimated, and was shown to compensate for missing data. The method of \cite{loh2011high} additionally has convergence guarantees for the optimization problem used to infer the parameters and can operate on data corrupted with multiplicative noise. 

In contrast to single and multivariate data, where one often has more control over the error in the independent variables, e.g. the placement of sensors, functional data originate from signals which are generally noisy, particularly when measured in extreme environments or at high sampling rates. This makes it essential that operator regression methods be robust to noise in both the input (independent) and output (dependent) variables. 
Error-in-variables models have previously been developed for linear function-to-function regression \citep{chakraborty2017regression,chen2022functional}. However, to the best of the authors' knowledge, there is no error-in-variables model available for any operator regression or nonlinear function-to-function regression methods. 

In this work, we derive a generalization to \eqref{eq:attenuation} for OLS inference of linear operators and demonstrate computationally that the action of OLS learned operators underpredict the action of the true operators. Under the assumption that the underlying functional data is smooth but corrupted by white noise, we propose EiV models for the nonlinear operator regression methods, MOR-Physics \citep{patel2018nonlinear,patel2021physics} and DeepONet \citep{lu2019deeponet}. We demonstrate that these models can correct for the bias introduced by noisy independent variables for several PDE learning problems.

\section{Operator regression for noisy input data}
Many operator regression methods seek to infer an operator by solving an ordinary least squares problem,
\begin{equation} \label{eq:lstsqinf}
    \mathcal{L} = 
\underset{\hat{\mathcal{L}}}{\mathrm{argmin}} \ \underset{(u,v)}{\mathbb{E}}
\left[\left\| \hat{\mathcal{L}} u - v \right\| _V^2 \right],
\end{equation}
where $\mathcal{L}$ is potentially nonlinear, $(u,v) \in U \times V$ represent the input and output of the target operator and the expectation $\mathbb{E}$ is over a distribution of functions that is appropriate for a given application, including any stochastic variables meant to model measurement noise. The specification of the Banach spaces $U$ and $V$ and the distribution inherent to the expectation \eqref{eq:lstsqinf} of input functions is a major theoretical challenge. The distribution weights a subset within the function space $U$ where the learned operator is expected to provide an accurate surrogate of the target operator, as well as measurement noise. In practice, it assumed that a finite sample of training functions is available which represents this theoretical distribution, and that these functions are discretized in some way so that they can be represented as finite-dimensional vectors,
\begin{equation}\label{eq:discretized_uv}
\mathbf{u} \in \mathbb{R}^{d_1},
\quad
\mathbf{v} \in \mathbb{R}^{d_2},
\end{equation}
It is also assumed that the operator $\mathcal{L}$ is discretized in a consistent way, 
\begin{equation}\label{eq:discretized_L}
\mathcal{L}: \mathbb{R}^{d_1} \rightarrow \mathbb{R}^{d_2},
\end{equation}
and that $\| \cdot \|_V 
= \| \cdot \|_{\ell_2(\mathbb{R}^{d_2})}$
for ease of computation.

We consider the case when the observations $(u,v)$ originate from pairs $(\hat{u}, \hat{v})$ corrupted by independent white noise. With the discretization specified by \eqref{eq:discretized_uv} and \eqref{eq:discretized_L}, and the analogous discretization of $(\hat{u}, \hat{v})$, we assume that we are given 
$N$  observations such that
\begin{align} \label{eq:data}
\begin{split}
\mathbf{u}^i
&=
\hat{\mathbf{u}}^i+\bm{\epsilon}_u^i,
\qquad \,
\mathbf{\mathbf{v}}^i =
\hat{\mathbf{v}}^i+\bm{\epsilon}_v^i,
\qquad
i = 1, 2, ..., N, 
\\
\bm{\epsilon}_{\mathbf{u}}^i 
&\sim \mathcal{N}(\mathbf{0},\sigma^2_u I),
\quad
\bm{\epsilon}_{\mathbf{v}}^i \sim \mathcal{N}(\mathbf{0},\sigma^2_v I),
\end{split}
\end{align}
where $\mathcal{N}(\bm{\mu}, \Sigma)$ is the multivariate normal distribution with mean $\bm{\mu}$ and covariance $\Sigma$. We seek an operator $ \mathcal{L}$ such that 
\begin{equation}
\mathcal{L}(\hat{\mathbf{u}}) \approx \hat{\mathbf{v}}
\end{equation}
for pairs of functions $(\hat{\mathbf{u}}, \hat{\mathbf{v}})$ in a given sample set (the test set). The discretized ordinary least squares problem from \eqref{eq:lstsqinf} for operator regression using this data is to minimize the loss
\begin{equation}\label{eq:lstsq_discretized}
J
=
\frac{1}{N}
\sum_{i=1}^N
\|
\mathcal{L} \mathbf{u}^i - \mathbf{v}^i
\|^2_{\ell_2(\mathbb{R}^{d_2})}.
\end{equation}
We stress that the objective is to infer the mapping from noiseless input to noiseless output, having access only to noisy observations of input and output data for training. 

\subsection{Operator regression methods}

In this section, we summarize the two operator regression methods examined in our experiments.

\subsubsection{MOR-Physics} \label{sec:morp}

The MOR-Physics operator regression method \citep{patel2018nonlinear,patel2021physics} uses the following operator parameterization,
\begin{equation}
\mathcal{L}(u) = \sum_i^{N_o} \mathcal{F}^{-1} g_i(\kappa) \mathcal{F} h_i(u),
\end{equation}
where $\mathcal{F}$ is the Fourier transform, $g_i$ and $h_i$ are neural networks, $\kappa$ is the wavenumber, and $N_o$ is a hyperparameter. Here, $h_i(u)$ is applied pointwise, such that, $h_i(u(x)) = (h_i\circ u)(x)$. The optimization over $\mathcal{L}$ in \eqref{eq:lstsqinf} is replaced here with optimization over the neural network parameters. For all numerical studies, we select the network $\text{width}=5$, $\text{depth}=5$, and take the activation function to be the exponential linear unit (ELU) \citep{clevert2015elu} for both $h_i$ and $g_i$.

\subsubsection{DeepONet}\label{sec:deeponet}

The DeepONet operator regression method \citep{lu2019deeponet} includes a branch network, $b$, that takes as input $u$ and a trunk network, $t$, that takes as input $x$, a single grid point on the output function $v$. In this work, we use the unstacked variant of DeepONet with the following operator parameterization to approximate $v$ at a single grid point $x$, 
\begin{equation}
    \mathcal{L}(u, x) = \sum_{k=1}^p b_k(u) t_k (x) + b_0,
\end{equation}
where $p$ is the network width and $b_0$ is an additional bias term.

For all computational studies involving DeepONet, we use fully connected networks with tanh activation functions for both the trunk and branch. We chose depths of 3 and 2 for the trunk and branch, respectively and 70 for the network width, $p$.  

\subsection{Bias in operators inferred from Ordinary Least-Squares (OLS) operator regression}

\begin{figure}[htpb]
    \centering
    \includegraphics[width=\textwidth]{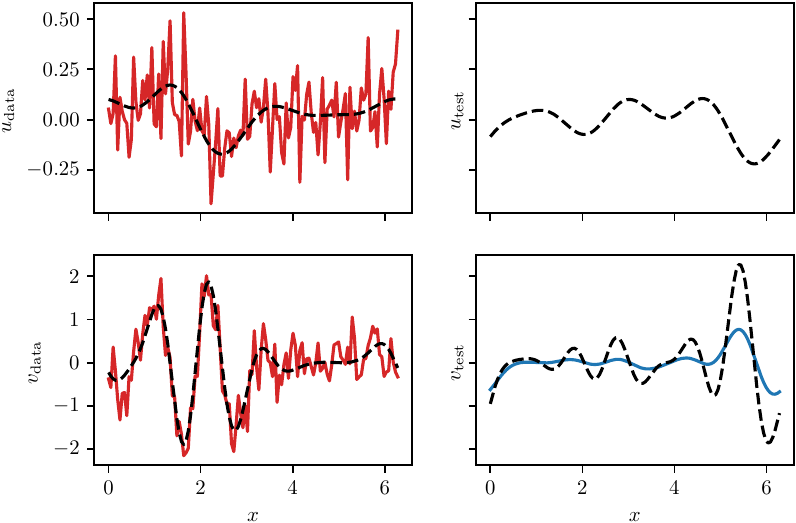}
    \caption{Operator learned from noisy independent variables with OLS underpredicts the action of the true operator. \textit{(Top left)} sample of training $u$ \sym{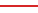} and underlying smooth function $\hat{u}$ \sym{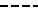}. \textit{(Bottom left)} sample of training $v$ \sym{C3,l,n,1} and underlying smooth function $\hat{v}=\partial_x\hat{u}^2$ \sym{k,d,n,1}. \textit{(Top right)} smooth test function. \textit{(Bottom right)} action of true Burgers operator on smooth test function \sym{k,d,n,1} and action of OLS learned operator \sym{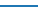}.} 
    \label{fig:bias}
\end{figure}

The presence of noise given by \eqref{eq:data} when minimizing a loss function as in \eqref{eq:lstsqinf} leads to systematic bias in the prediction of $\mathcal{L}$, in a way that is analogous to the attenuation bias described previously.
We generate pairs of noisy functions,
\begin{equation}
 (u^i,v^i) = (\hat{u}^i + \epsilon_u^i,\partial_x \hat{u}^{i^2} + \epsilon_v^i),
\end{equation}
where $u^i$ and $v^i$ are generated as described in Section \ref{sec:results},
and attempt to infer the Burgers operator $ \mathcal{L}u = \partial_x u^2$ using the MOR-Physics operator regression method, described in Section \ref{sec:morp}. 
Figure~\ref{fig:bias} demonstrates the bias in the learned operator by plotting the action on a test function; the predicted output function exhibits $\ell_2$-norm smaller than the true output function. This is typical of the output on test functions, suggesting that the operator norm of $\mathcal{L}$ itself is attenuated, i.e., that the OLS learned operator ``underpredicts'' the action of the true Burgers operator. In Table~\ref{tab:OLSstats}, we compute the maximum and average norms for the action of the OLS learned and true Burgers operator over a set of 1000 test functions in the unit ball of $\ell_2$; note that the former approximates the operator norm of the learned $\mathcal{L}$ over the sample space. These statistics strongly suggest that 
the operator norm is biased to zero.

\begin{table}[h!]
\centering
\begin{center} 
\begin{tabular}{ c | c | c }
   & OLS operator & True operator \\
    \hline 
    $\max \|\mathcal{L} u \|_{\ell_2(\mathbb{R}^{d_2})}$                        & 4.42 & 12.45 \\ 
    $\mathrm{average}\, \|\mathcal{L} u \|_{\ell_2(\mathbb{R}^{d_2})}$            & 2.40 & 7.75 
\end{tabular}
\end{center}
\caption{Operator norm statistics for true Burgers operator  and  OLS inferred operator, computed over 1000 samples with $|| u ||_{\ell_2(\mathbb{R}^{d_1})}=1$.}
\label{tab:OLSstats}
\end{table}

\begin{figure}[htpb]
    \centering
    \includegraphics[width=\textwidth]{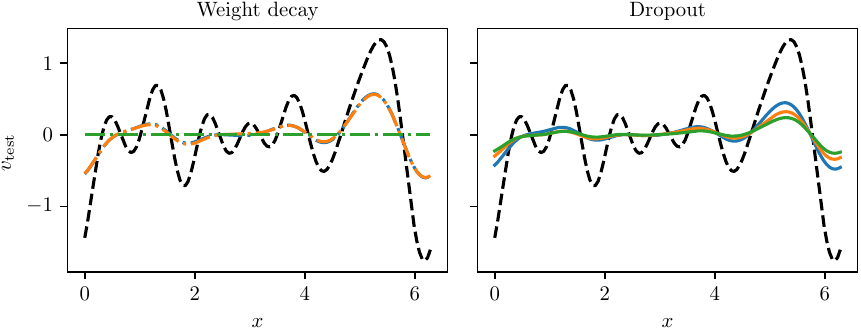}
    \caption{Standard neural network regularization methods do not reduce bias in operators learned from noisy independent variables. Action of true Burgers operator \sym{k,d,n,1}. \textit{(Left)} Weight decay with penalties, $\lambda =0.01$ \sym{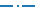},  $\lambda = 0.1$ \sym{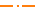}, and $\lambda =1$ \sym{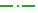}. \textit{(Right)} Dropout with dropout rates, $r = 0.01$ \sym{C0,l,n,1},  $r=0.05$ \sym{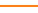}, and $r=0.1$ \sym{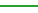}. }
    \label{fig:bias_reg}
\end{figure}

In addition to the bias induced by noise in the independent variable, a natural question is whether overfitting by the DNN is a cause of attenuation bias. 
In Figure~\ref{fig:bias_reg}, we again attempt to recover the Burgers operator but include regularization techniques for neural networks to prevent overfitting, i.e., weight decay \citep{krogh1991simple} and dropout \citep{srivastava2014dropout}. For various penalties and dropout rates, the OLS learned operators still underpredict the action of the true Burgers operator. This suggests that the observed bias is not a consequence of overfitting, but a result of inadequate error modelling. The hyperparameters used in these studies are available in Appendix~\ref{sec:hyp}.

The following theorem shows that the bias observed in the examples above can be expected for linear operators. It extends the classical result on attenuation bias to the case of a linear operator learned by OLS. We obtain the theorem for an idealized case of infinitely many finite-dimensional samples. That is, we assume a discretization of the form \eqref{eq:discretized_uv}, but for infinitely many pairs, consistent with the implicit distribution of samples. Then, rather than the fully discrete loss \eqref{eq:lstsq_discretized}, the theorem is stated for a loss of the form
\begin{equation}\label{eq:intermediate_loss}
\overline{J}
=
{\mathbb{E}}
\left[\left\| \hat{\mathcal{L}} \mathbf{u} - \mathbf{v} \right\|^2 _{\ell_2(\mathbb{R}^{d_2})} \right],
\quad
\mathbf{u} \in \mathbb{R}^{d_1},
\mathbf{v} \in \mathbb{R}^{d_2}.
\end{equation}
The fully discrete loss given in \eqref{eq:lstsq_discretized}
converges to $\overline{J}$ as the sample size increases. The following theorem is proven in Appendix \ref{sec:appendix_proof}.

\begin{theorem}\label{thm:attenuation_bias}
The minimizer  to \eqref{eq:intermediate_loss} among linear $\hat{\mathcal{L}}$ is
\begin{equation}
\mathcal{L}
=
\mathbb{E}
\left[
\hat{\mathbf{v}} \hat{\mathbf{u}}^\top
\right]
\left(
\mathbb{E}
\left[
\hat{\mathbf{u}} \hat{\mathbf{u}}^\top
\right]
+
\sigma^2_u I
\right)^{-1}.
\end{equation}
The norm upperbound for this operator is
\begin{equation}
\| \mathcal{L} \| \leq 
\frac{\| \mathbb{E}
\left[
\hat{\mathbf{v}} \hat{\mathbf{u}}^\top
\right] \|}
{ \|
\mathbb{E}
\left[
\hat{\mathbf{u}} \hat{\mathbf{u}}^\top
\right]
+
\sigma^2_u I
\| }.
\end{equation}
\end{theorem}
This norm upperbound decreases with increasing levels of noise, $\sigma_u$, suggesting attenuation bias for the OLS inferred operator with large noise in $u$. Theorem \ref{thm:attenuation_bias} should be compared to the attenuation bias result for classical regression \eqref{eq:attenuation}. The significance of this result is that while zero-bias noise in the dependent variable $v$ can be compensated for by a large sample size, the presence of seemingly innocuous zero-bias noise in the independent variable $u$ persists in biasing the optimal solution of the ordinary least squares problem \eqref{eq:intermediate_loss} itself. The numerical results shown in Figure \ref{fig:bias} and Table \ref{tab:OLSstats} above suggests similar behavior for nonlinear operator regression as well. This necessitates the error-in-variables model we propose below, which corrects the bias in the predicted operator. 


%

\section{Error-in-variables (EiV) model} \label{sec:eiv}


Instead of applying OLS to infer the operator, we model the error for both $u$ and $v$ with the joint probability density function (PDF),
\begin{equation}
    \begin{bmatrix}
        {\tilde{u}^i} - {u}^i\\
        {v^i} - \mathcal{L}\tilde{u}^i
    \end{bmatrix}
    \sim
    \mathcal{N}\left(\mathbf{0},
        \begin{bmatrix}
            \sigma^2_u I &  \\
              & \sigma^2_v I
        \end{bmatrix}
    \right)
\end{equation}
%
where $\tilde{u}$ is a ``denoised'' version of ${u}$ that ideally approximates the true underlying, noiseless function, $\hat{u}$. $\mathcal{L}$, along with $\tilde{u}, \sigma_u, \sigma_v$, can be recovered via maximum likelihood estimation (MLE),
\begin{equation}\label{eq:opt}
    \mathcal{L},\tilde{u},\sigma_u,\sigma_v = \underset{\mathcal{L}, \tilde{u}, \sigma_u, \sigma_v}{\mathrm{argmax}} \prod_i 
    p\left(\begin{bmatrix}
        \tilde{u}^i - {u}^i\\
        v^i - \mathcal{L} \tilde{u}^i
    \end{bmatrix}\right),
\end{equation}
where $p(\cdot)$ refers to the probability density of the indicated random variable. 
Real-world functional data is often smooth. Therefore, we simplify recovery of $\tilde{u}$ via a low-pass filter, $\tilde{u} = \mathcal{G}u$, and optimize over $\mathcal{G}$ instead of $\tilde{u}$. We select for $\mathcal{G}$, the smooth spectral filter \citep{boyd1996erfc}, 
\begin{equation}
    \mathcal{G}(u) = \mathcal{F}^{-1}\mathrm{erfc}(a(\kappa-\kappa_c))\mathcal{F}u
\end{equation}
where $\mathcal{F}$, $a$, $\kappa$, and $\kappa_c$ are the Fourier transform, the filter bandwidth, the wavenumber, and the cutoff wavenumber, respectively. Since $u$ is assumed to be smooth and all examples in Section~\ref{sec:results} are performed on the periodic domain, this spectral filter as a simple and efficient denoising method. The optimization over $\tilde{u} = \mathcal{G}u$ in \eqref{eq:opt} is replaced with optimization over the two filter parameters.

In our numerical studies, we have observed that priors on the filter parameters can improve robustness. Ideally, we would like as weak a prior as possible on $\kappa_c$, since we do not \textit{a priori} know the smoothness of $u$. A uniform prior would be the weakest, but its PDF is discontinuous near the boundaries of the support. Instead, we use the Beta distribution \citep{murphy2013machine} as a prior on $\kappa_c$,
\begin{equation}
\begin{aligned}
    \kappa_c/\beta_{\kappa_c} \sim \mathrm{Beta}(1+\varepsilon,1+\varepsilon).
\end{aligned}
\end{equation}
where $\beta_{\kappa_c}$ is a hyperparameter that specifies the maximum allowable $\kappa_c$. With this prior, we perform maximum a posteriori (MAP) estimation to obtain $\mathcal{L}$,
\begin{equation}\label{eq:optmap}
    \mathcal{L},\kappa_c,a,\sigma_u,\sigma_v = \underset{\mathcal{L}, \kappa_c, a, \sigma_u, \sigma_v}{\mathrm{argmax}} \prod_i 
    p\left(\begin{bmatrix}
            \mathcal{G}u^i - {u}^i\\
        v^i - \mathcal{L} \mathcal{G}{u}^i
\end{bmatrix}\right) p(\kappa_c/\beta_{\kappa_c}). 
\end{equation}
For small $\varepsilon$, $\mathrm{Beta}$ is a smooth approximation to the uniform distribution, and therefore MAP is amenable to gradient descent based algorithms. This prior forces our EiV model to select a smaller $\kappa_c$ so that $\mathcal{G}u^i$ is smooth.  For our numerical studies, we use $\varepsilon=0.01$.



\section{Operator regression with error-in-variables model for time-evolving systems}\label{sec:time}

In addition to the inference problem discussed in the previous sections, we are also interested in inferring time-evolving PDE's from noise corrupted solutions. Given a solution, $u(x,t) = \hat{u}(x,t) + \epsilon(x,t)$, we seek to recover a PDE in the following form,
\begin{align}
    \partial_t u &= \mathcal{L}(u), \quad x \in \Omega \\
    u(x,0) &= u_0(x) \\
    \mathcal{B}u &= 0, \quad x \in \partial \Omega
\end{align}
We discretize this PDE in time with forward Euler, obtaining,
\begin{equation}
    \begin{aligned}
        u_{n+1} = u_n + \Delta t \mathcal{L}(u_n) = \mathcal{P}(u_n) \quad n=1,2,\hdots,\hat{N}_t,
    \end{aligned}
\end{equation}
This temporal discretization results in a ResNet-type architecture which has been effectively used to parameterize and discover governing evolution equations \citep{Haber2017,Long2018,patel2018nonlinear,qin2019data}. We still parameterize $\mathcal{L}$ with either MOR-Physics or DeepONet. However, we can now ideally model the error in $u$ over all time steps with the following joint PDF,
\begin{equation}
\begin{bmatrix}
\mathcal{G} u_0 - u_0 \\
\mathcal{P} \mathcal{G} u_0 -u_1 \\
\vdots \\
\mathcal{P}^{\hat{N}_t} \mathcal{G} u_0 -u_n \\
\end{bmatrix}
\sim
\mathcal{N}\left(\mathbf{0},
\sigma_u^2 I 
\right).\end{equation}
Note that we have assumed the error is spatially and temporally independent and identical distributed (i.i.d.). The validity of this assumption depends on the data and the physical system under investigation.
Due to the computational complexity of evaluating probability densities from the above distribution, we instead consider the following marginalized distribution as a simplification,
\begin{equation}
\begin{bmatrix}
\mathcal{G} u_0 - u_0 \\
\mathcal{P}^{N_t} \mathcal{G} u_0 -u_{N_t}
\end{bmatrix}
\sim
\mathcal{N}\left(\mathbf{0},
\sigma^2_u I 
\right),
\end{equation}
where $N_t$ is a hyperparameter indicating a maximum timestep.

\section{Results}\label{sec:results}

In the following sections, we compare OLS and EiV operator regression using both MOR-Physics and DeepONet. We perform all test cases on the periodic domain, $\Omega = [0,L]^d$, where $d$ is the spatial dimension of the problem. For each test, we generate samples of smooth functions, $\hat{\mathbf{u}}^i$ by applying a low pass filter to samples of white noise.  For time independent problems, we then apply the \textit{a priori} known operator to $\hat{\mathbf{u}}^i$ and obtain $\hat{\mathbf{v}}^i$.  

We next corrupt the data by adding white noise and obtain $\mathbf{u}^i$ and $\mathbf{v}^i$. To standardize the amount of noise added to $\hat{\mathbf{u}}$ and $\hat{\mathbf{v}}$, we tune the standard deviation of the white noise to target specific signal-to-noise ratios (SNR's), expressed in decibels (dB) as,
\begin{equation}
    \mathrm{SNR} = 10 \log_{10} \left(\frac{\mathrm{RMS}(\mathbf{u})}{\sigma_u}\right)^2,
\end{equation}
and likewise for $\mathbf{v}$. For all test cases, we target the same SNR for both $\mathbf{u}$ and $\mathbf{v}$. With this noisy data, we attempt to infer the operator or PDE. Finally we evaluate our inference by generating noiseless test data, $u_{\mathrm{test}}$ using the same method discussed above, and comparing the action of the inferred operator to the action of the true operator on $u_{\mathrm{test}}$.

For time evolving problems, we numerically integrate the \textit{a priori} known PDE, using $\hat{\mathbf{u}}^i_0 = \hat{\mathbf{u}}^i$ as the initial condition, obtaining $\hat{\mathbf{u}}^i_n$ for all $n$ timesteps. We next corrupt this data with white noise as discussed above, obtaining $\mathbf{u}_n$, from which we attempt to infer an evolution PDE. We evaluate our inference by generating a noiseless initial condition $\mathbf{u}_0=\mathbf{u}_{\mathrm{test}}$, evolving the inferred PDE, and comparing to the evolution of the true PDE using the same $\mathbf{u}_{\mathrm{test}}$ as an initial condition.

To perform the optimization in \eqref{eq:opt} and \eqref{eq:optmap}, we use the stochastic gradient based optimizer, ADAM \citep{Kingma2014}. Batch sizes, learning rates, sample sizes, and other hyperparameters for all numerical studies are listed in Appendix~\ref{sec:hyp}.

\subsection{Learning operators with the MOR-Physics EiV model}

In these sections, we compare the OLS and EiV models using the MOR-Physics operator parameterization discussed in Section~\ref{sec:morp}.

\subsubsection{Recovering the Burgers operator}

\begin{figure}[t]
    \centering
    \includegraphics[width=\textwidth]{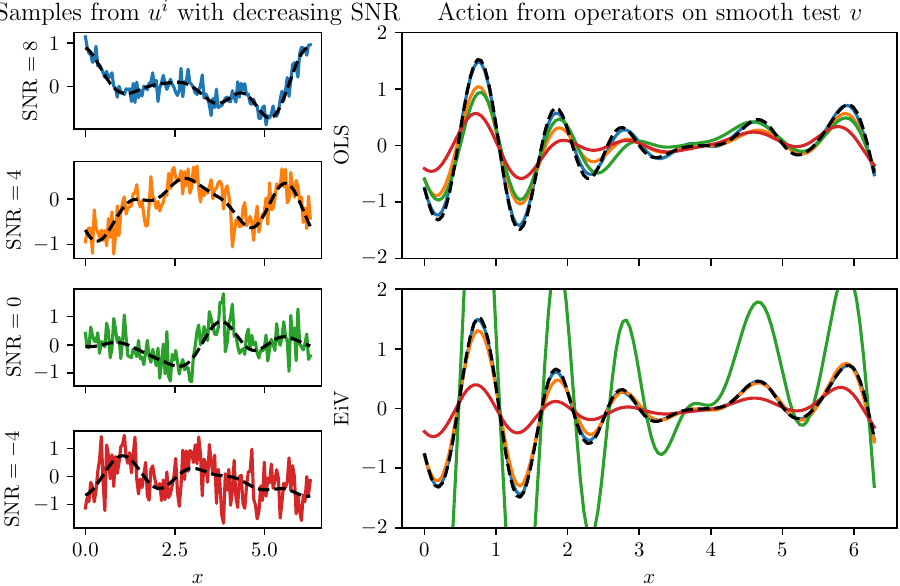}
    \caption{ EiV model improves recovery of true Burgers operator in the presence of noisy independent variables. \textit{(Left)} Underlying smooth function $\hat{u}$ \sym{k,d,n,1} and training $u$ for $\mathrm{SNR} = 8$ \sym{C0,l,n,1}, $\mathrm{SNR} = 4$ \sym{C1,l,n,1}, $\mathrm{SNR} = 0$ \sym{C2,l,n,1}, and $\mathrm{SNR} = -4$ \sym{C3,l,n,1}. \textit{(Right)} Action of true Burgers operator \sym{k,d,n,1} on noiseless test $\mathbf{u}_{\mathrm{test}}$ and action of learned operators from data with decreasing SNR for OLS \textit{(Top right)} and EiV \textit{(Bottom right)}.}
    \label{fig:burgers_eiv}
\end{figure}

We consider the learning problem discussed in Figure~\ref{fig:bias} where we attempt to recover the Burgers operator, $\mathcal{L}u=\partial_xu^2$. In Figure~\ref{fig:burgers_eiv}, we apply OLS and EiV inferred error models to this dataset, recover operators, compute the action of these operators on a noiseless test function, $u_{\mathrm{test}}$, and compare to the action of the true operator, $v_{\mathrm{test}}$. We observe that the OLS model recovered operators only reproduce the action of the true Burgers operator for the highest SNR examined, while underpredicting the action for lower SNR's. In contrast, the EiV model is able to recover operators down to $\mathrm{SNR}=4$ with high accuracy.  However, for lower SNR's the EiV model also fails to recover a suitable operator.

\begin{figure}[htpb]
    \centering
    \includegraphics[width=0.7\textwidth]{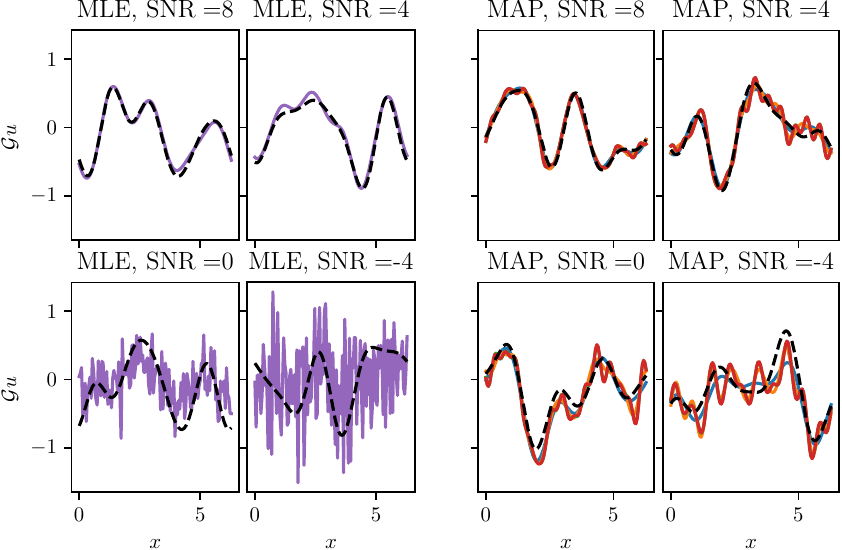}
    \caption{ Effect of cutoff wavenumber prior on filter for EiV model. \textit{(Left)} Action of MLE estimate of filters on noisy $u^i$ \sym{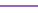} for decreasing SNR and corresponding noiseless $\hat{u}^i$ \sym{k,d,n,1}. \textit{(Right)} Action of MAP estimate of filters ($\kappa_c$ prior) on $u^i$ with hyperparameters, $\beta_{\kappa_c}=10$ \sym{C0,l,n,1}, $\beta_{\kappa_c}=20 \sym{C1,l,n,1}$, $\beta_{\kappa_c}=40 \sym{C2,l,n,1}$, and $\beta_{\kappa_c}=80 \sym{C3,l,n,1}$.}
    \label{fig:filt}
\end{figure}

On the left four subplots of Figure~\ref{fig:filt}, we examine the action of the filter $\mathcal{G}$ on samples of noisy input functions, $u^i$, with decreasing SNR. For the high SNR data, the EiV model successfully identifies filter parameters that smooth the input functions. However, for low SNR data, the EiV model selects filters that fail to smooth the input functions, suggesting the method finds high $\kappa_c$.

\subsubsection{Improved operator learning with smoothness prior}

\begin{figure}[htpb]
    \centering
    \includegraphics[width=0.7\textwidth]{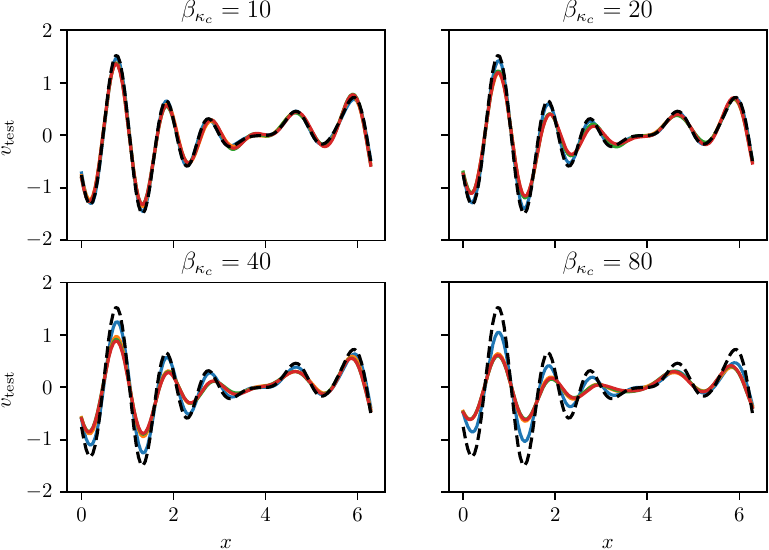}
    \caption{Cutoff wavenumber prior improves EiV model. Action of EiV operator on $u_{\mathrm{test}}$ learned from $\mathrm{SNR}=8$ \sym{C0,l,n,1}, $\mathrm{SNR}=4$ \sym{C1,l,n,1}, $\mathrm{SNR}=0$ \sym{C2,l,n,1}, and $\mathrm{SNR}=-4$ \sym{C3,l,n,1} for various $\beta_{\kappa_c}$. Action of true operator \sym{k,d,n,1}.}
    \label{fig:prior}
    
\end{figure}

The results in Figure~\ref{fig:burgers_eiv} can be improved by adding priors on the filter parameters as discussed in Section~\ref{sec:eiv} to constrain $\kappa_c$ to smaller values and enforce smoothness on $\mathcal{G}u^i$. In Figure~\ref{fig:prior}, we plot the action of the learned operators with priors on $\kappa_c$ and four different hyperparameters, $\beta_{\kappa_c}$. We find, for a wide range of $\beta_{\kappa_{c}}$, that the prior enables successful recovery of the Burgers operator even for the lowest SNR. However for the largest $\beta_{\kappa_c}=80$, our EiV model can no longer consistently recover the Burgers operator. Notably, these results are relatively insensitive to $\beta_{\kappa_c}$, so careful hyperparameter selection is not necessary.

On the right four subplots of Figure~\ref{fig:filt}, we examine the action of the filter $\mathcal{G}$ on sample noisy input functions, $u^i$, with decreasing SNR and the four $\beta_{\kappa_c}$ hyperparameters. We find that the MAP estimate produces an operative filter that smooths $u^i$, even for the lowest SNR.

\subsubsection{Learning the 2D Burgers operator}

\begin{figure}[t]
    \centering
    \includegraphics[width=0.8\textwidth]{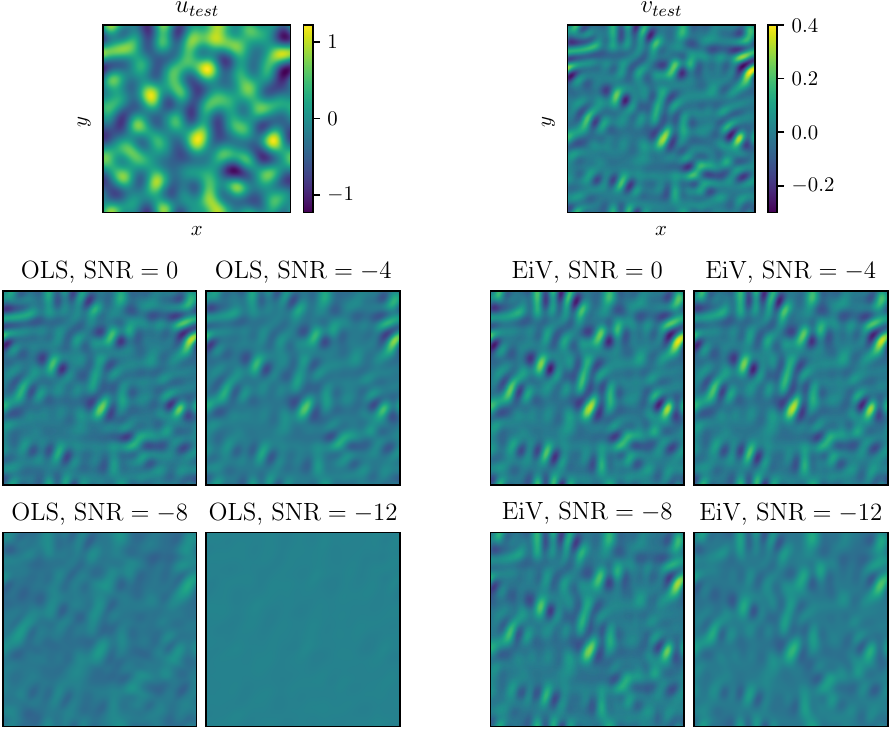}
    \caption{EiV model improves recovery of 2D Burgers operator from noisy data. \textit{(Top)} Noiseless test data, $u_{\mathrm{test}}$, and action of true operator, $v_{\mathrm{test}}$.  Action of OLS \textit{(Bottom left)}  and EiV \textit{(Bottom right)} learned operators on $u_{\mathrm{test}}$ for decreasing SNR.}
    \label{fig:2d}
\end{figure}

In this section, we attempt to recover a 2D generalization of the Burgers operator \citep{Mohamed2019}, $\mathcal{L} = \partial_xu^2 + \partial_yu^2$. Figure~\ref{fig:2d} compares the OLS and EiV learned operators for decreasing SNR. For the EiV model, we included the $\kappa_c$ prior with $\beta_{\kappa_c}=10$. As in the 1D Burgers operator test, we find that action of the OLS learned operator severely underpredicts the action of the true operator, especially at very low SNR. The EiV learned operator, however, captured the action of the true operator with SNR down to $-4$. Even with lower SNR's, the EiV operator did not underpredict the true operator as severely as the OLS operator.

\subsubsection{Learning the Kuramoto–Sivashinsky equation}

\begin{figure}[t]
    \centering
    \includegraphics[width=0.8\textwidth]{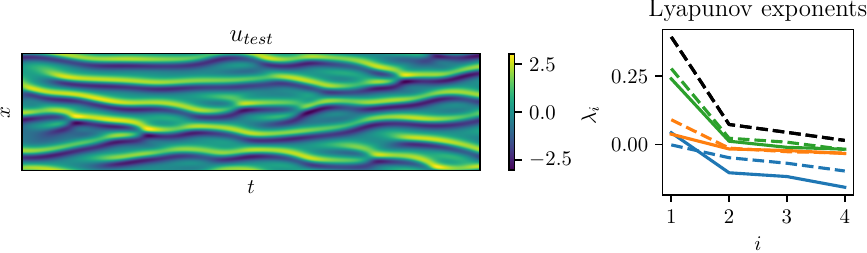}
    \includegraphics[width=0.8\textwidth]{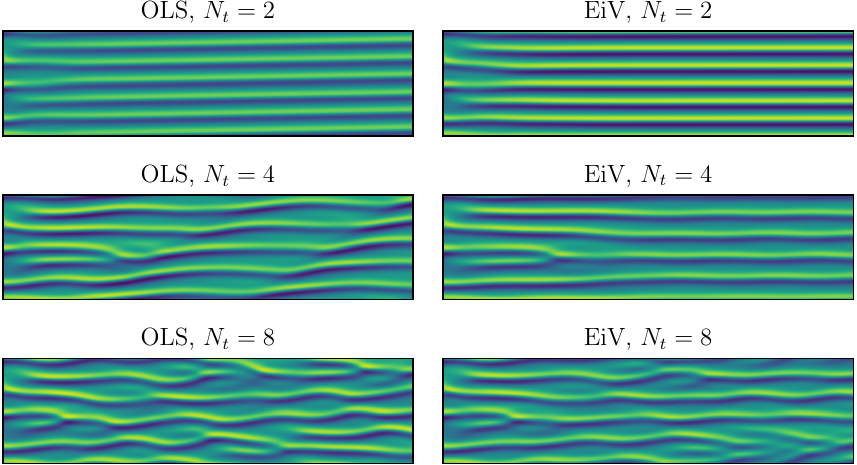}
    \caption{OLS and EiV models perform similarly for KS equation inference. \textit{(Top left)} Noiseless test data, $u_{\mathrm{test}}$. \textit{(Bottom left)} OLS and \textit{(Bottom right)} EiV inferred operators for increasing hyperparameter, $N_t$. \textit{(Top right)} Lyapunov exponents for true equation \sym{k,d,n,1}; OLS equation with $N_t=2$ \sym{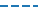}, $N_t=4$ \sym{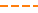}, $N_t=8$ \sym{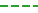}; and EiV equation with $N_t=2$ \sym{C0,l,n,1}, $N_t=4$ \sym{C1,l,n,1}, $N_t=8$ \sym{C2,l,n,1}.}
    \label{fig:ks}
\end{figure}

In this section we attempt to recover the Kuramoto–Sivashinsky (KS) equation,
\begin{equation}
    \partial_t u + 0.5 \partial_x u^2 + \partial_x^2u + \partial_x^4u = 0,
\end{equation}
using the method outlined in Section~\ref{sec:time}. To generate the data, we solve KS equation using Pseudospectral methods. We use Orszag's 3/2 zero-padding technique to eliminate aliases from the Fourier transform of $u^2$, which is computed in the ($x$,$t$) domain \citep{Orszag1971}. We use the low-storage Runge-Kutta method for temporal discretization \citep{Spalart1991} which has been widely used in the direct numerical simulation of turbulent flows \citep{Hoyas2006,Lee2015}. It treats the nonlinear term explicitly with third-order accuracy. Linear terms are implicitly treated, similarly to the Crank-Nicolson method, with second-order accuracy.

In Figure~\ref{fig:ks}, we infer the equation using the OLS and EiV models for increasing final timestep, $N_t$, and found that both performed similarly. For this study, we used $\mathrm{SNR}=4$. Qualitatively, neither model was able to capture the dynamics of the KS equation with $N_t=2$ and $N_t=4$, but both models were able to capture the dynamics with $N_t=8$.

As a chaotic system, the KS equation can be characterized by computing Lyapunov exponents as outlined in \cite{edson2019}. In Figure~\ref{fig:ks}, we also compute the first four Lyapunov exponents for the true and learned equations. As observed in the qualitative comparison, OLS and EiV perform similarly, only successfully recovering the true KS equation for $N_t=8$. 
This is in contrast to regressing the Burgers operator, in which EiV outperformed OLS. We hypothesize that is due to the difference in regularity of the target operators; the Burgers operator, a differential operator, reduces the regularity of the input function, while the KS evolution operator possesses smoothing properties \citep{collet1993analyticity}. As a result, one can expect the operator norm over most subspaces of sample functions to be much higher for the Burgers operator than for the KS evolution operator, rendering the OLS attenuation bias more significant in regressing the former. 
We plan to explore in more detail the advantages of EiV over OLS in time dependent PDE learning and time independent operator learning in future work.

\subsection{Learning the Burgers operator with DeepONet EiV model}

\begin{figure}[tb]
    \centering
    \includegraphics[width=0.7\textwidth]{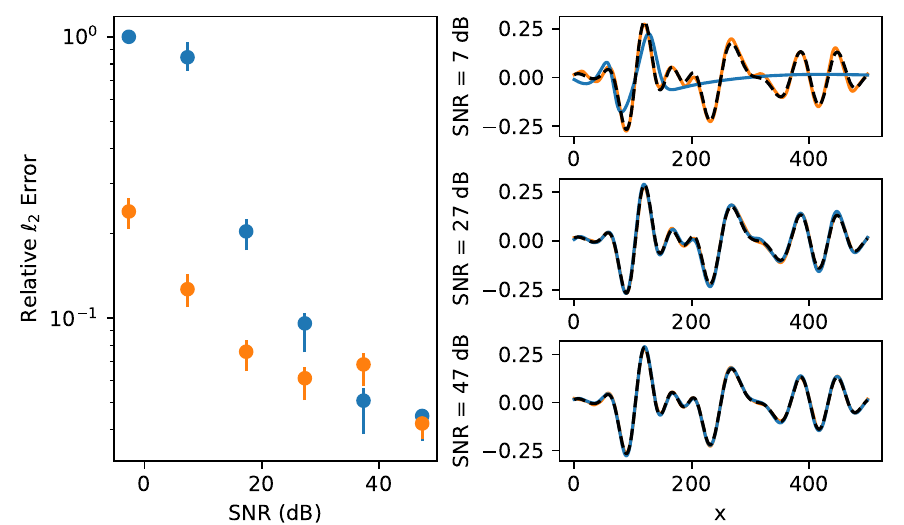}
    \caption{EiV inference performs better than OLS inference for low SNR in DeepONet. \textit{(Left)} Relative $\ell_2$ error vs. SNR for OLS \sym{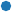} and EiV \sym{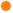}. \textit{(Right)} Action of true \sym{k,d,n,1}, OLS \sym{C0,l,n,1}, and EiV \sym{C1,l,n,1} operators on $u_{\mathrm{test}}$ for increasing SNR. Error bars indicate 25th and 75th percentile over 100 test samples.}
    \label{fig:snr}
\end{figure}

\begin{figure}[htpb]
    \centering
    \includegraphics[width=0.8\textwidth]{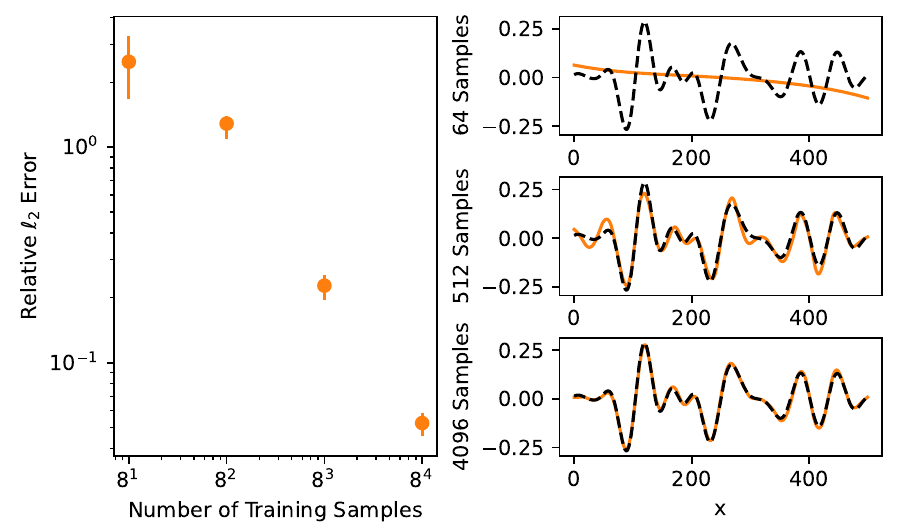}
    \caption{EiV inference improves with training set size in DeepONet. \textit{(Left)} Relative $\ell_2$ error vs. training set size for EiV \sym{C1,n,o,1}. \textit{(Right)} Action of true \sym{k,d,n,1} and EiV \sym{C1,l,n,1} operators on $u_{\mathrm{test}}$ for increasing training set size. Results computed for $\mathrm{SNR}=17$. Error bars indicate 25th and 75th percentile over 100 test samples.}
    \label{fig:train}
\end{figure}

In this section, we examine EiV operator learning for the 1D Burgers operator using the DeepONet method outlined in Section~\ref{sec:deeponet}. In Figure~\ref{fig:snr}, we compute the $\ell_2$ error between the actions of OLS and EiV learned operators on test functions and the true actions. As in MOR-Physics, the EiV learned operator robustly learns from noisy data, even at very low SNR. Finally, we examine the effect of training set size on EiV learning in Figure~\ref{fig:train}. We find that for very low samples, EiV fails to recover the Burgers operator to high accuracy. With larger numbers of samples, EiV can successfully recover the operator. 

\begin{figure}[htpb]
    \centering
    \includegraphics[width=0.8\textwidth]{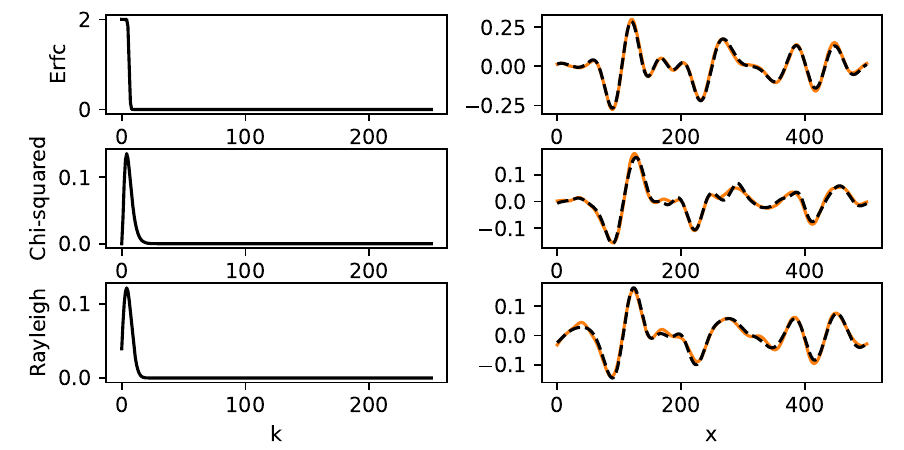}
    \caption{Effect of spectral filter used to generate input signals. \textit{(Left)} Frequency content of $u_{\mathrm{test}}$. \textit{(Right)} Action of true \sym{k,d,n,1} and DeepONet EiV \sym{C1,l,n,1} operators on $u_{\mathrm{test}}$.} 
    \label{fig:freqspec}
\end{figure}

To test the robustness of our EiV model, we next consider alternative distributions from which to sample the smooth functions $u$. To create the dataset we repeat the procedure discussed in Section~\ref{sec:results}, except instead of low pass filtering white noise to generate $u$, we apply one of the following Fourier kernels to white noise,
\begin{equation}
\begin{aligned}
    &K(\kappa) = \mathrm{erfc(\kappa - 6}) \\
    &K(\kappa) = \frac{1}{2} \kappa^{2} e^{-\kappa/2} \\
    &K(\kappa) = \frac{\kappa-1}{25}e^{-(\kappa-1)^2/50}
    \end{aligned}
\end{equation}
where the second and third functions are PDFs of the $\chi^2$ distribution with 6 degrees of freedom and the Rayleigh distribution with location parameter $=-1$ and scale parameter $=5$. With the transfer function the smooth input functions are computed as 
\begin{equation}
\begin{aligned}
    &u(x) = \mathcal{F} K(\kappa) e^{-jR(\kappa)}\\
    &R(\kappa) \sim \mathcal{U}[0,2\pi]
\end{aligned}
\end{equation}
where $\mathcal{U}$ is the uniform distribution and $j$ is the imaginary number. These transfer functions are chosen such that $u$ is smooth and has no high frequency content. We find in Figure~\ref{fig:freqspec} that  DeepONet with our EiV model is able to successfully recover the Burgers operator regardless of the distribution used to generate $u$.

\section{Conclusion}
We have demonstrated that operator regression performed by minimizing a least-squares loss is prone to attenuation bias when the input, or independent variable, is corrupted by white noise. This is supported both by an upper bound on the operator norm for linear operator regression using ordinary least squares (OLS) as well as numerical results for nonlinear operator regression for problems such as recovering the Burgers operator from noisy observations. We proposed an Error-in-Variables (EiV) model for operator regression that replaces the least-squares loss with maximum likelihood estimation or maximum a posteriori estimation with an appropriate smoothness prior. The EiV method is applicable to a wide variety of existing operator regression methods; we demonstrate this by combining it with the MOR-physics method of \citet{patel2018nonlinear} and the DeepONet method of \cite{lu2019deeponet}. For these methods, the EiV model significantly improves the operator prediction fidelity for a variety of problems given noise in the input variable, as demonstrated for the Burgers equation in one and two dimensions. Finally, we introduce an EiV model for time evolving PDEs and show OLS and EiV perform similarly in recovering the Kuramoto-Sivashinky equation. This suggests that the regularity properties of the target operator effect the severity of OLS attenuation bias in practice. For many physical systems, the white noise assumption in this work may be invalid. Future work may consider other types of error, such as multiplicative noise and spatially correlated noise. We may also examine alternative means of denoising, e.g., total variation denoising \citep{vogel1996iterative}, and alternative operator learning frameworks, e.g., wavelet neural operators \citep{gupta2021multiwaveletbased}. Additionally, while we have solely focused on obtaining MAP estimates for operators, particularly in small data regimes, the posterior distributions may hold other likely operators. Future work may focus on examining this posterior distribution of operators.

\acks{Sandia National Laboratories is a multimission laboratory managed and operated by National Technology and Engineering Solutions of Sandia, LLC, a wholly owned subsidiary of Honeywell International, Inc., for the U.S. Department of Energy’s National Nuclear Security Administration under contract DE-NA0003525. This paper describes objective technical results and analysis. Any subjective views or opinions that might be expressed in the paper do not necessarily represent the views of the U.S. Department of Energy or the United States Government.

The work of R. Patel, I. Manickam, and M. Gulian has also been supported by the U.S. Department of Energy, Office of Advanced Scientific Computing Research under the Collaboratory on Mathematics and Physics-Informed Learning Machines for Multiscale and Multiphysics Problems (PhILMs) project. SAND Number: SAND2022-5179 C.}

\bibliography{ref.bib}

\appendix

\section{Proof of Theorem 1}\label{sec:appendix_proof}
We write the objective function in \eqref{eq:intermediate_loss} as
\begin{align}
\overline{J} &= 
\mathbb{E} 
\left[ (\mathcal{L} \mathbf{u} - \mathbf{v})^\top (\mathcal{L} \mathbf{u} - \mathbf{v}) \right]
\\
&=
\mathbb{E}
\left[
(\mathbf{u}^\top \mathcal{L}^\top - \mathbf{v}^\top) (\mathcal{L} \mathbf{u} - \mathbf{v}) \right] 
\\
&=
\mathbb{E}
\left[ \mathbf{u}^\top \mathcal{L}^\top \mathcal{L} \mathbf{u} - \mathbf{v}^\top \mathcal{L} \mathbf{u} - \mathbf{u}^\top \mathcal{L}^\top \mathbf{v} + \mathbf{v}^\top \mathbf{v} \right]
\\
&=
\mathbb{E}
\left[ \mathbf{u}^\top \mathcal{L}^\top \mathcal{L} \mathbf{u} - 2 \mathbf{v}^\top \mathcal{L} \mathbf{u} + \mathbf{v}^\top \mathbf{v} \right].
\end{align}
Using linearity of expectation, this can be written as a sum of three terms, namely
\begin{align}
I&=
\mathbb{E}
\left[ \mathbf{u}^\top \mathcal{L}^\top \mathcal{L} \mathbf{u} \right]
\\
&=
\mathbb{E}
\left[ (\hat{\mathbf{u}} + \bm{\epsilon}_u)^\top 
\mathcal{L}^\top \mathcal{L} 
(\hat{\mathbf{u}} + \bm{\epsilon}_u) \right]
\\
&=
\mathbb{E}
\left[ (\hat{\mathbf{u}}^\top + \bm{\epsilon}_u^\top)
\mathcal{L}^\top \mathcal{L} 
(\hat{\mathbf{u}} + \bm{\epsilon}_u) \right]
\\
&=
\mathbb{E}
\left[ 
\hat{\mathbf{u}}^\top
\mathcal{L}^\top \mathcal{L} 
\hat{\mathbf{u}}
+
\hat{\mathbf{u}}^\top
\mathcal{L}^\top \mathcal{L} 
\bm{\epsilon}_u
+
\bm{\epsilon}_u^\top
\mathcal{L}^\top \mathcal{L} 
\hat{\mathbf{u}}
+
\bm{\epsilon}_u^\top
\mathcal{L}^\top \mathcal{L} 
\bm{\epsilon}_u
\right]
\\
&=
\mathbb{E}
\left[ 
\hat{\mathbf{u}}^\top
\mathcal{L}^\top \mathcal{L} 
\hat{\mathbf{u}}
+
2
\bm{\epsilon}_u^\top
\mathcal{L}^\top \mathcal{L} 
\hat{\mathbf{u}}
+
\bm{\epsilon}_u^\top
\mathcal{L}^\top \mathcal{L} 
\bm{\epsilon}_u
\right]
\\
&=
\mathbb{E}
\left[ 
\hat{\mathbf{u}}^\top
\mathcal{L}^\top \mathcal{L} 
\hat{\mathbf{u}}
\right]
+
\mathbb{E}
\left[ 
\bm{\epsilon}_u^\top
\mathcal{L}^\top \mathcal{L} 
\bm{\epsilon}_u
\right]
;
\end{align}
\begin{align}
II
&=
-2
\mathbb{E}
\left[
\mathbf{v}^\top \mathcal{L} \mathbf{u} 
\right]
\\
 &=
 -2
\mathbb{E}
\left[
(\hat{\mathbf{v}}+\bm{\epsilon}_v)^\top 
\mathcal{L}
(\hat{\mathbf{u}} + \bm{\epsilon}_u)
\right]
 \\
 &=
 -2
\mathbb{E}
\left[
\hat{\mathbf{v}}^\top 
\mathcal{L}
\hat{\mathbf{u}}
+
\hat{\mathbf{v}}^\top
\mathcal{L}
\bm{\epsilon}_u
+
\bm{\epsilon}_v^\top 
\mathcal{L}
\hat{\mathbf{u}}
+
\bm{\epsilon}_v^\top 
\mathcal{L}
 \bm{\epsilon}_u
\right]
\\
&=
 -2
\left(
 \mathbb{E}
\left[ 
\hat{\mathbf{v}}^\top 
\mathcal{L}
\hat{\mathbf{u}}
\right]
+
\hat{\mathbf{v}}^\top
\mathcal{L}
\left[
\mathbb{E}
\bm{\epsilon}_u
\right]
+
\left[
\mathbb{E}
\bm{\epsilon}_v^\top
\right]
\mathcal{L}
\hat{\mathbf{u}}
+
\left[
\mathbb{E}
\bm{\epsilon}_v^\top 
\right]
\mathcal{L}
\left[
\mathbb{E}
 \bm{\epsilon}_u
\right]
 \right)
\\
 &=
 -2
 \mathbb{E}
\left[ 
\hat{\mathbf{v}}^\top 
\mathcal{L}
\hat{\mathbf{u}}
\right]
;
\end{align}
and 
$
III = 
\mathbb{E}
\left[ \mathbf{v}^\top \mathbf{v}
\right].
$
While $\partial III/\partial \mathcal{L} = 0$, we have
\begin{align}
\frac{\partial I}{\partial \mathcal{L}} &= 
\mathbb{E}
\left[ 
\frac{\partial }{\partial \mathcal{L}}
\hat{\mathbf{u}}^\top
\mathcal{L}^\top \mathcal{L} 
\hat{\mathbf{u}}
\right]
+
\mathbb{E}
\left[ 
\frac{\partial }{\partial \mathcal{L}}
\bm{\epsilon}_u^\top
\mathcal{L}^\top \mathcal{L} 
\bm{\epsilon}_u
\right]
\\
&=
\mathbb{E}
\left[ 
2 \mathcal{L} \hat{\mathbf{u}} \hat{\mathbf{u}}^\top 
\right]
+
\mathbb{E}
\left[
2 \mathcal{L}
\bm{\epsilon}_u \bm{\epsilon}_u^\top
\right]
\\
&=
2 \mathcal{L}
\mathbb{E}
\left[  \hat{\mathbf{u}} \hat{\mathbf{u}}^\top 
\right]
+
2 \mathcal{L}
\mathbb{E}
\left[
\bm{\epsilon}_u \bm{\epsilon}_u^\top
\right]
\\
&=
2\mathcal{L} \left(
\mathbb{E}
\left[
\hat{\mathbf{u}} \hat{\mathbf{u}}^\top
\right]
+
\sigma^2_u I
\right),
\end{align}
and
\begin{align}
\frac{\partial II}{\partial \mathcal{L}}
&= 
-2
\mathbb{E}
\left[
\frac{\partial }{\partial \mathcal{L}}
\hat{\mathbf{v}}^\top \mathcal{L} \hat{\mathbf{u}}
\right]
\\
&= 
-2
\mathbb{E}
\left[
\hat{\mathbf{v}} \hat{\mathbf{u}}^\top
\right]
.
\end{align}
Setting the derivative of the objective function equal to zero, we obtain
\begin{equation}
\mathcal{L} \left(
\mathbb{E}
\left[
\hat{\mathbf{u}} \hat{\mathbf{u}}^\top
\right]
+
\sigma^2_u I
\right)
= 
\mathbb{E}
\left[
\hat{\mathbf{v}} \hat{\mathbf{u}}^\top
\right]
\end{equation}
which has solution
\begin{equation}
\mathcal{L}
=
\mathbb{E}
\left[
\hat{\mathbf{v}} \hat{\mathbf{u}}^\top
\right]
\left(
\mathbb{E}
\left[
\hat{\mathbf{u}} \hat{\mathbf{u}}^\top
\right]
+
\sigma^2_u I
\right)^{-1}.
\end{equation}

\newpage
\section{Hyperparameters}\label{sec:hyp}

The following tables lists the hyperparameters used in our numerical examples. The multiple learning rates indicate that we reinitialized the Adam optimizer during training and lowered the learning rate.

\begin{table}[h!]
\centering
\begin{center} 
\begin{tabular}{ c | c | c | c | c | c  }
   & Figures~\ref{fig:bias}, \ref{fig:bias_reg}, \ref{fig:burgers_eiv}, \ref{fig:filt}, and \ref{fig:prior}  & Figure~\ref{fig:2d}  & Figure~\ref{fig:ks}  & Figures~\ref{fig:snr} and \ref{fig:train} & Figure~\ref{fig:freqspec}\\
    \hline 
    Domain size, $L$                 & $2\pi$   &  $2\pi \times 2\pi$ & $30\pi$ & 512   & 512               \\
    Grid size                        & 128      &  $128 \times 128$   & 256     & 512   & 512        \\
    Training set size                & 64       &  $64$               & 128     & 1500  & 3000       \\
    Batch size                       & 4        &  $4$                & 32      & 1500  & 3000       \\
    Learning rate 1                  & 1E-3     &  1E-3               & 1E-3    & 1e-4  & 1e-3      \\
    Learning rate 2                  & 1E-4     &  1E-4               & 1E-3    & 5e-5  & 3e-4         \\
    Learning rate 3                  & 1E-5     &  N/A                & 1E-3    & 1e-5  & 5e-4      \\
    Epochs                           & 400      &  200                & 1       & 1.5E6 & 1.5E6     \\
    Number of operators, $N_o$       & 1        &  1                  & 2       & N/A   & N/A  
\end{tabular}
\end{center}
\caption{Hyperparameters used for all computational studies.}
\label{tab:hyp}
\end{table}

\end{document}